\pdfoutput=1

\documentclass[11pt]{article}
\usepackage{authblk}
\usepackage{acl}
\usepackage{enumitem}
\usepackage{array}
\usepackage{makecell}
\usepackage{times}
\usepackage{latexsym}
\usepackage{booktabs}
\usepackage{graphicx} 
\usepackage{multirow}
\usepackage[T1]{fontenc}

\usepackage[utf8]{inputenc}

\usepackage{microtype}

\usepackage{inconsolata}
\usepackage{latexsym}
\usepackage{microtype}
\usepackage{hyperref}
\usepackage{xcolor}
\usepackage[misc]{ifsym}
\title{SceMQA: A Scientific College Entrance Level Multimodal Question Answering Benchmark}

\author[1]{\textbf{Zhenwen Liang}}
\author[1]{\textbf{Kehan Guo}}
\author[1]{\textbf{Gang Liu}}
\author[1]{\textbf{Taicheng Guo}}
\author[1]{\textbf{Yujun Zhou}}
\author[1]{\textbf{Tianyu Yang}}
\author[2]{\\ \textbf{Jiajun Jiao}}
\author[3]{\textbf{Renjie Pi}}
\author[3]{\textbf{Jipeng Zhang}}
\author[1]{\textbf{Xiangliang Zhang} \textsuperscript{\tiny \Letter}}
\affil[1]{University of Notre Dame, \texttt{\{zliang6,xzhang33\}@nd.edu}}
\affil[2]{New York University} 
\affil[3]{Hong Kong University of Science and Technology}

\begin{document}
\maketitle
\begin{abstract}
The paper introduces SceMQA, a novel benchmark for scientific multimodal question answering at the college entrance level. It addresses a critical educational phase often overlooked in existing benchmarks, spanning high school to pre-college levels. SceMQA focuses on core science subjects including Mathematics, Physics, Chemistry, and Biology. It features a blend of multiple-choice and free-response formats, ensuring a comprehensive evaluation of AI models' abilities. Additionally, our benchmark provides specific knowledge points for each problem and detailed explanations for each answer. SceMQA also uniquely presents problems with identical contexts but varied questions to facilitate a more thorough and accurate assessment of reasoning capabilities. In the experiment, we evaluate both open-source and close-source state-of-the-art Multimodal Large Language Models (MLLMs), across various experimental settings. The results show that further research and development are needed in developing more capable MLLM, as highlighted by only 50\% to 60\% accuracy achieved by the strongest models. Our benchmark and analysis will be available at \url{https://scemqa.github.io/}.
\end{abstract}

\begin{figure}
\centering 
\includegraphics[width=0.46\textwidth]{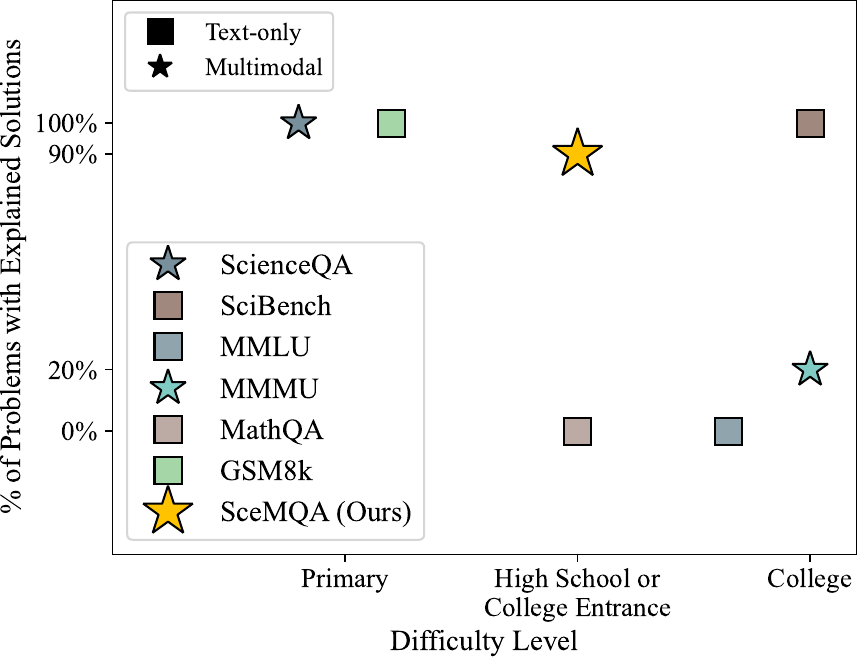} 
\caption{The comparison between SceMQA and other existing benchmarks. Y-axis is the percentage of problems that have detailed solution explanations. Most problems (over 90\%) in SceMQA has detailed explanations to solutions except for some straightforward problems. More comparison can be found in Table \ref{tab:benchmark_comparison}.} 
\label{fig:intro} 
\end{figure}

\section{Introduction}
In recent years, the evolution of large language models (LLMs) has marked a significant milestone in artificial intelligence. Initially, these models excelled in diverse natural language processing tasks \cite{brown2020language,ouyang2022training,touvron2023llama1,touvron2023llama2,openai2023gpt4,google2023gemini}, but their utility has since increasingly expanded, transforming them into incredible agents for various downstream tasks such as reasoning and planning \cite{li2023camel,wu2023autogen,park2023generative,guolarge}. Notably, LLMs have shown proficiency in tasks that typically pose significant challenges to even highly skilled humans, such as tackling intricate mathematical problems \cite{lu2023mathvista,romera2023mathematical} and accelerating scientific discoveries \cite{birhane2023science}. This evolution demonstrates the versatility of LLMs and their potential to revolutionize areas traditionally dominated by human expertise.

\begin{figure*}
\centering 
\includegraphics[width=1.00\textwidth]{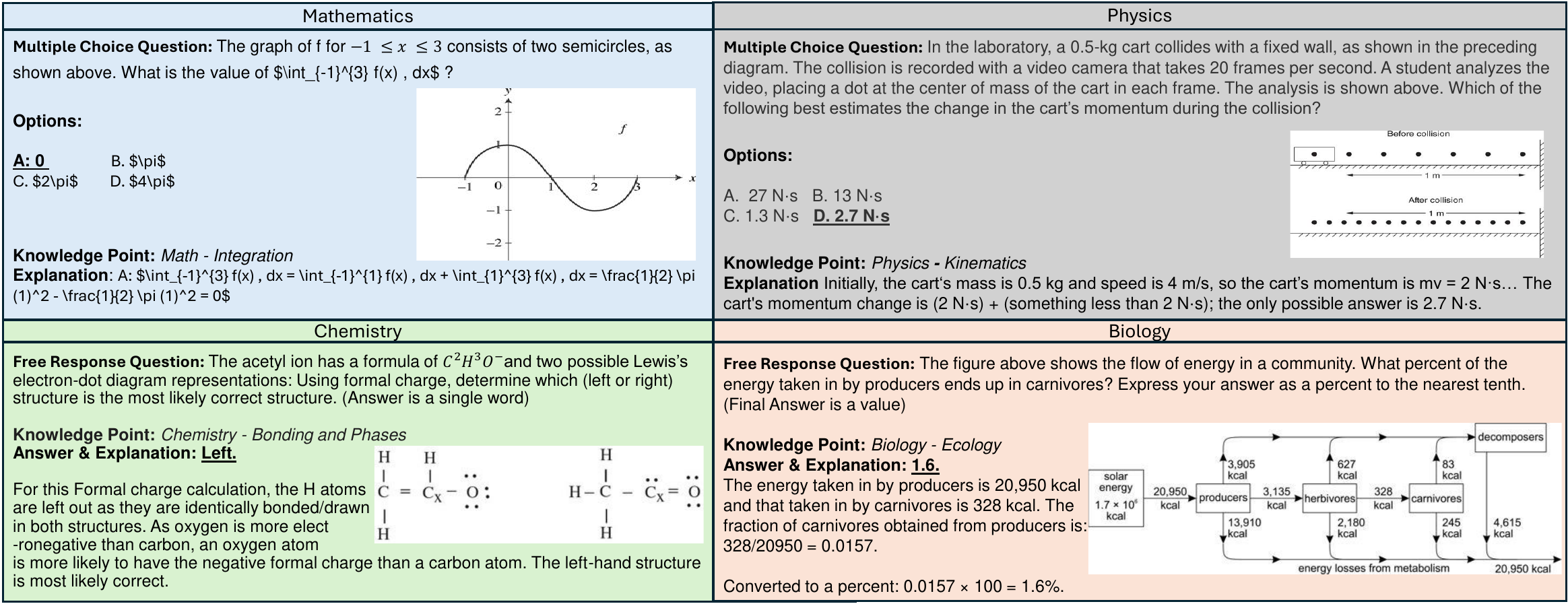} 
\caption{Example problems in SceMQA, which contains four scientific subjects - math, physics, chemistry and biology in two formats - multiple choice and free response.}
\label{fig:examples} 
\end{figure*}
Alongside, the rapid development of vision-based LLMs has garnered considerable attention within the AI community, especially with the release of platforms like OpenAI's GPT4-V \cite{openai2023gpt4v} and Google's Gemini Ultra \cite{google2023gemini}. These models have demonstrated exceptional abilities in tasks requiring advanced reasoning and planning, often surpassing existing benchmarks and approaching human-level performance. This progress has spurred researchers to create more sophisticated and challenging benchmarks for Multimodal LLMs (MLLMs), one of the most representative is the science domain, which is a long-standing focus for humans. For example, the MathVista benchmark \cite{lu2023mathvista}, comprising 6,141 problems, demands a high level of visual understanding and mathematical reasoning. Additionally, the Massive Multi-discipline Multimodal Understanding and Reasoning Benchmark (MMMU) \cite{yue2023mmmu} poses college-level multimodal reasoning challenges. Currently, even the most advanced models achieve only about 50\% accuracy on these benchmarks. The importance of such benchmarks lies in their role as vital tools for assessing and pushing the boundaries of AI capabilities. By presenting AI models with tasks that mimic complex, real-world scenarios, benchmarks provide a clear measure of progress and highlight areas for future development.

However, in the science domain, a critical observation in multimodal reasoning benchmarks is the disparity in the levels of difficulty. Prior benchmarks like ScienceQA \cite{lu2022learn} primarily focused on elementary and middle-school levels, while MMMU leaps to a college-level challenge. This leaves a significant educational phase in human learning – the high school, or college entrance level – relatively unaddressed.  In fact, learning progressively in difficulty levels is not only important for humans, but also can facilitate AI systems including LLMs via curriculum learning \cite{bengio2009curriculum} and progressive training \cite{xu2023contrastive,mitra2023orca}. Therefore, we fill this gap by introducing a novel benchmark named \textbf{S}cience \textbf{c}ollege \textbf{e}ntrance level \textbf{M}ultimodal \textbf{Q}uestion \textbf{A}nswering (SceMQA), designed for this critical educational stage, encompassing four key subjects: Mathematics \& Statistics, Physics, Chemistry, and Biology. 

Apart from the difficulty level, our benchmark also has a detailed annotation granularity. Firstly, SceMQA includes detailed, human-verified explanations for each answer. Besides, each problem is classified into a specific knowledge component, facilitating detailed knowledge tracing for models. Moreover, SceMQA uniquely features problems with the same context but different questions. This design is informed by prior research indicating that without diverse question types for each narrative context, models might resort to learning shallow heuristics or patterns rather than developing a deep, semantic understanding \cite{patel2021nlp,yang-etal-2022-unbiased}. This approach ensures a more comprehensive and precise evaluation of reasoning capabilities.
In Figure \ref{fig:intro}, we compare the difficulty level, annotation granularity, and covered modality among existing benchmarks. Example problems in our benchmark are shown in Figure \ref{fig:examples}.

\section{Related Work}

\paragraph{Multimodal Question Answering}
Multimodal Question Answering (QA) has been a focal area in AI research. The Visual Question Answering (VQA) benchmark \cite{antol2015vqa}, established in 2015, pioneered free-form, open-ended visual QA, necessitating intricate image comprehension and reasoning. ChartQA\cite{masry2022chartqa} emphasized complex reasoning about charts, merging visual and logical thought processes. VisIT-Bench \cite{bitton2023visit} tested vision-language models across real-world tasks, ranging from simple recognition to advanced creative generation.
\begin{table*}[htbp]
\small
\centering
\resizebox{2.05\columnwidth}{!}{
\begin{tabular}{lccccc}
\toprule
 &  Problem Format & \# Problems Per Subject & Problem Modality & Solution Explanation* & Difficulty Level \\
\midrule
MMLU & MC & 279 & T & No & College \\
SciBench & FR & 232 & T & Yes & College \\
ScienceQA & MC & 816 & T+I & Yes & Primary \\
MathVista & MC + FR & - & T+I & No & Unspecified \\
MMMU & MC + FR & 385 & T+I & No & College \\
SceMQA (Ours) & \textbf{MC + FR} & 261 & \textbf{T+I} & \textbf{Yes} & \textbf{College Entrance} \\
\bottomrule
\end{tabular}}
\caption{A comparative overview of various benchmarks. The first column indicates the problem types inside the benchmark, with ``MC'' representing multiple choice and ``FR'' indicating free-response formats. The second column shows the average number of problems per subject. The third column describes the problem modality, where ``I'' stands for image-based and ``T'' for text-based problems. (*) The fourth column categorizes benchmarks based on whether over 90\% of problems are annotated with solutions explanations. The final column presents the difficulty level. All superior and unique features of our benchmark are highlighted.}
\label{tab:benchmark_comparison}
\end{table*}
\paragraph{Multimodal LLMs}
In addition to notable models like GPT4-V and Google Gemini, various open-source Multimodal LLMs (MLLMs) have emerged. MiniGPT-4 \cite{zhu2023minigpt} improved vision-language understanding by syncing a visual encoder with a language LLM. LLaVAR \cite{zhang2023llavar} combined OCR with text-only GPT-4 for enhanced visual instruction tuning in text-rich image contexts. mPLUG-Owl \cite{ye2023mplug} proposed a modular framework for equipping LLMs with multimodal capabilities, focusing on image-text alignment. InstructBLIP \cite{dai2023instructblip} excelled in vision-language instruction tuning, demonstrating remarkable zero-shot performance in diverse tasks. For a more detailed summary of related studies, please refer to these surveys \cite{wu2023multimodal,yin2023survey}.

\paragraph{Science Question Answering}
Various benchmarks have been developed for specific scientific subjects, including  MATH \cite{hendrycks2021measuring}, MathVista \cite{lu2023mathvista}, chemistry \cite{guo2023indeed}, etc. More comprehensive science QA benchmarks like ScienceQA \cite{lu2022learn}, C-EVAL \cite{huang2023c}, AGIEVAL \cite{zhong2023agieval}, MMMU \cite{yue2023mmmu}, and SciBench \cite{wang2023scibench} have recently been introduced, providing a broader scope of assessment.

{In this paper, propose a benchmark SceMQA to fill the gap of multi-modal science QA at the  college entrance level.}
A comparative analysis of our dataset against existing benchmarks is detailed in Table \ref{tab:benchmark_comparison}. Although our benchmark appears smaller in total problem count, it focuses specifically on the science domain, offering a substantial average number of problems per subject. Furthermore, it excels in quality, as evidenced by the high proportion of problems accompanied by detailed explanations.

\section{Our Benchmark SceMQA}

\subsection{Overview of SceMQA}

\begin{figure}
\centering 
\includegraphics[width=0.48\textwidth]{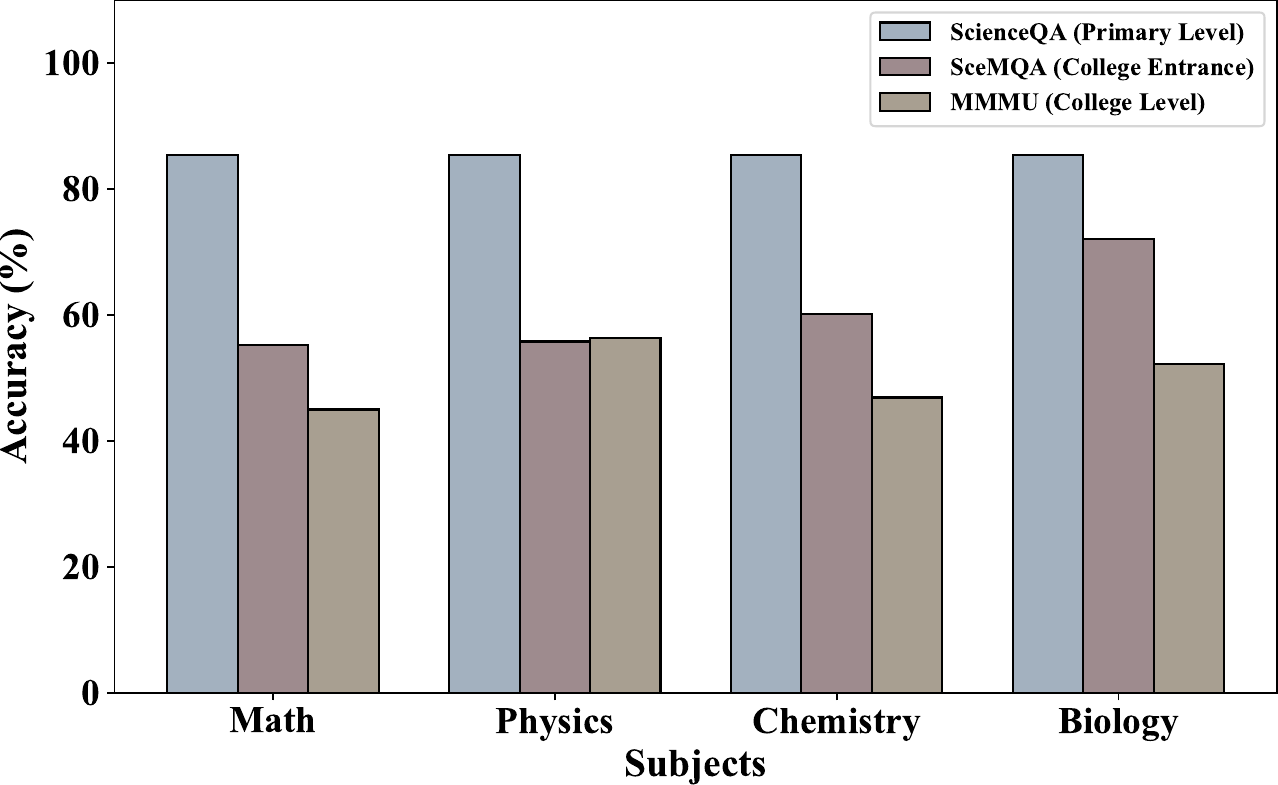} 
\caption{Comparison of GPT-4 performance across different benchmarks, illustrating the accuracy percentages achieved by GPT-4 in different subjects.} 
\label{fig:compare} 
\end{figure}

Our benchmark is strategically designed to bridge a significant gap in existing multimodal benchmarks, which typically span from elementary to college levels, overlooking the crucial high school and pre-college stages. This educational phase is crucial in the human learning process. Although some existing benchmarks \cite{zhong2023agieval,zhang2023evaluating} incorporate problems at this level, they predominantly feature text-only questions. 
Our benchmark stands out as the first college entrance level multimodal benchmark within the research community, offering a more comprehensive assessment tool. 
To evaluate the difficulty of the problems in our benchmark, we utilize GPT-4 to respond to the questions within our dataset, as well as those from both a primary level and a college level benchmark. 
Figure \ref{fig:compare} demonstrates the  moderate difficulty level  of our benchmark, positioning between the existing benchmark on primary and college levels. 

\paragraph{Science Subjects} Focusing on the core science subjects such as mathematics, physics, biology, and chemistry, our benchmark aligns with both existing text-only benchmarks, such as SciBench \cite{wang2023scibench}, and major human exams like the GaoKao (i.e., Chinese national college entrance exam). To effectively address these problems, AI models must demonstrate a robust understanding of images, tables, and diagrams, coupled with deep domain knowledge to recall necessary formulae, theorems, and other elements for advanced reasoning. This presents a suitable challenge for current AI systems, testing their limits in areas typically reserved for advanced human cognition.

\paragraph{Solution Explanation} We have meticulously annotated every problem in SceMQA. Almost all solutions (> 90\%) are accompanied by detailed, human-verified explanations except for some straightforward solutions, as shown in Figure \ref{fig:examples}. These explanations are useful for indentifying errors in model predictions and could also be instrumental in future supervised fine-tuning (SFT) \cite{ho2022large,hsieh2023distilling} and few-shot prompting methodologies \cite{wei2022chain}. 

\paragraph{Identified Knowledge Category} Additionally, each problem is associated with  specific knowledge components within its subject, also shown in Figure \ref{fig:examples}. The availability of these components aids in building a knowledge state for the evaluated models, facilitating knowledge tracing and understanding the depth of the model's capabilities. 

\begin{figure}
\centering 
\includegraphics[width=0.48\textwidth]{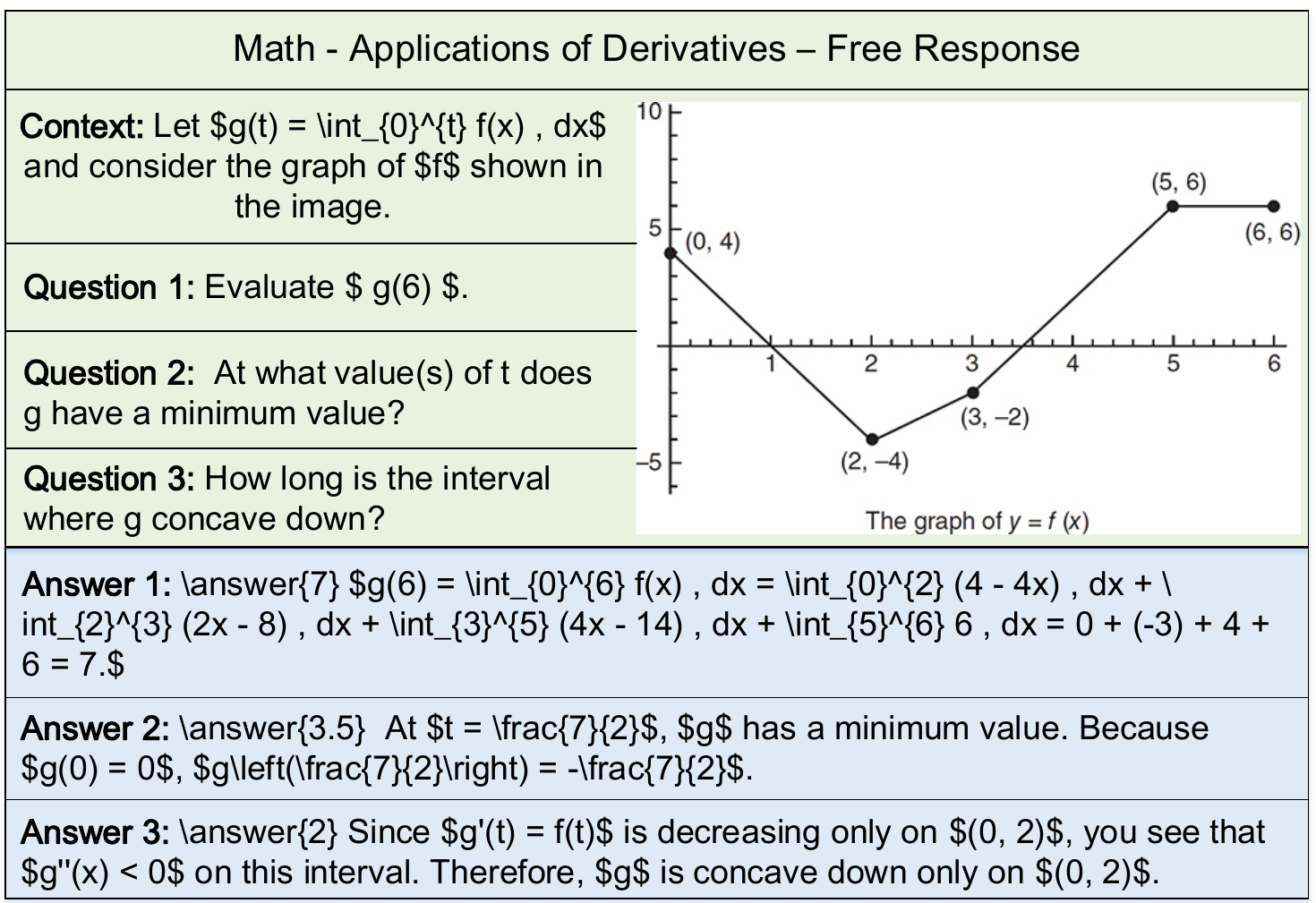} 
\caption{SceMQA contains multiple questions under the same context.} 
\label{fig:variation} 
\end{figure}

\paragraph{Question Variation} Furthermore, our benchmark features a variety of questions based on the same image and context, as shown in Figure \ref{fig:variation}. Solving such kind of question sets has been demonstrated to be challenging for AI models \cite{liang2021solving}, where they usually fail to detect subtle differences among various questions related to the same context \cite{patel2021nlp}. This one-context multiple-questions  setting can not only test the depth of understanding and reasoning capabilities of these AI models \cite{patel2021nlp, yang-etal-2022-unbiased} but also have the potential to support advancements in Socratic learning \cite{shridhar-etal-2022-automatic} and interpretable reasoning \cite{zhang2021noahqa}.

\begin{table}[ht]
\small
\centering
\renewcommand{\arraystretch}{1.2} 
\begin{tabular}{l|>{\centering\arraybackslash}m{1.5cm}|>{\centering\arraybackslash}m{1.3cm}}
\hline
& \textbf{Multiple Choice} &  \textbf{Free Response}\\
\hline
Total Questions & 845 & 200 \\
Unique Images & 632 & 118 \\
Max Question Length & 1816 & 1906 \\
Max Answer Length & 1124 & 2614 \\
Average Question Length & 452 & 410\\
Average Answer Length & 297 & 330\\
\hline
\end{tabular}
\caption{SceMQA Statistics.}
\label{table:benchmark_statistics}
\end{table}

SceMQA has in total 1,045 problems,  
with an average of 261 problems per subject. Details can be found in Table \ref{table:benchmark_statistics}. This set of problems ensures a thorough evaluation across all included subjects.

\begin{table*}[h]
\centering
\renewcommand{\arraystretch}{1.2}
\resizebox{\textwidth}{!}{%
\begin{tabular}{|c|c|c|c|c|c|c|c|c|c|c|c|c|}
\hline
\multicolumn{12}{|c|}{Open-sourced models} \\ \hline
\multicolumn{2}{|c|}{\multirow{2}{*}{Model}} & \multicolumn{5}{c|}{Multiple Choice} & \multicolumn{5}{c|}{Free Response} \\ \cline{3-12}
\multicolumn{2}{|c|}{} & Math  & Physics  & Chemistry  & Biology & Overall  & Math  & Physics & Chemistry  & Biology  & Overall  \\ \hline
\multicolumn{2}{|c|}{InstructBLIP-7B}  & 16.98 & 21.86 & 20.30 & 22.75 & 20.48 & 6.00 & 6.00 & 0.00 & 38.00 & 12.50 \\ \hline
\multicolumn{2}{|c|}{InstructBLIP-13B}  & 19.34 & 19.53 & 17.33 & 28.91 & 21.31 & 8.00 & 12.00 & 4.00 & 30.00 & 13.50 \\ \hline
\multicolumn{2}{|c|}{MiniGPT4-7B}  & 18.87 & 20.93 & 25.25 & 22.75 & 21.90 & 4.00 & 0.00 & 2.00 & 20.00 & 6.50 \\ \hline
\multicolumn{2}{|c|}{MiniGPT4-13B}  & 27.39 & 20.93 & 27.23 & 35.55 & 27.74 & 2.00 & 4.00 & 8.00 & 14.00 & 7.00 \\ \hline
\multicolumn{2}{|c|}{LLaVA1.5-7B}  & 25.94 & 25.12 & 21.78 & 36.97 & 27.50 & 10.00 & 4.00 & 2.00 & 26.00 & 10.50 \\ \hline
\multicolumn{2}{|c|}{LLaVA1.5-13B}  & 31.13 & 28.37 & 26.24 & 38.86 & 31.19 & 12.00 & 4.00 & 4.00 & 32.00 &13.00 \\ \hline
\multicolumn{12}{|c|}{Close-sourced models} \\ \hline
\multirow{3}{*}[1.2ex]{Model} & \multirow{3}{*}[1.2ex]{Setting} & \multicolumn{5}{c|}{Multiple Choice} & \multicolumn{5}{c|}{Free Response} \\ \cline{3-12}
 & & Math  & Physics  & Chemistry  & Biology & Overall  & Math  & Physics & Chemistry  & Biology  & Overall  \\ \hline
 \multirow{1}{*}{Google Bard} & Text-only & 43.40 & 40.93 & 24.75 & 54.88 & 41.31 & - & - & - & - & - \\ \hline
\multirow{3}{*}{Gemini Pro} & Text-only & 21.70 & 19.53 & 32.51 & 46.51 & 30.06 & 8.00 & 6.00 & 8.00 & 38.00 & 15.00 \\ \cline{2-12}
& Few-shot & 36.79 & 30.23 & 37.44 & 48.84 & 38.34 & 18.00 & 12.00 & 12.00 & 36.00 & 19.50 \\ \cline{2-12}
& Zero-shot & 37.26 & 30.70 & 42.36 & 54.42 & 41.18 & 20.00 & 12.00 & 18.00 & 36.00 & 21.50 \\ \hline
\multirow{3}{*}{GPT4-V} & Text-only & 35.38 & 47.91 & 58.13 & 63.72 & 51.24 & 12.00 & 24.00 & 28.00 & 22.00 & 21.50 \\ \cline{2-12}
& Few-shot & 54.72 & 53.95 & 58.62 & 67.44 & 58.70 & 30.00 & 24.00 & 30.00 & 48.00 & 33.00 \\ \cline{2-12}
& Zero-shot & \textbf{55.19} & \textbf{55.81} & \textbf{60.10} & \textbf{72.09} & \textbf{60.83} & \textbf{36.00} & \textbf{24.00} & \textbf{36.00} & \textbf{48.00} & \textbf{36.00} \\ \hline
\end{tabular}%
}
\caption{Accuracy  of examining GPT4-V and Gemini Pro across different settings on Multiple Choice and Free Response problems in SceMQA.}
\label{tab:combined_results}
\end{table*}
\subsection{Data Collection Protocol}
The data for SceMQA was meticulously sourced from publicly available online materials tailored for college entrance level tests in four key subjects: math (including calculus and statistics), biology, physics, and chemistry. In selecting these questions, our team of annotators strictly adhered to the licensing regulations of the source websites, ensuring no copyrighted material was included. This adherence to legal and ethical standards was a priority throughout the data collection process.

Each problem within our dataset contains one image that is essential for solving the corresponding question, aligning with the multimodal nature of SceMQA. The problems are presented in two formats: multiple-choice and free-response. The multiple-choice questions offer 4 to 5 options, denoted by uppercase letters, a format consistent with other established benchmarks. Following previous studies \cite{hendryckstest2021,lewkowycz2022solving}, we transform all mathematical expressions into latex codes, making them easy to process for LLMs, as shown in Figure \ref{fig:examples} and \ref{fig:variation}.

The free-response section includes calculation-based problems where answers are numerical values. This format is particularly advantageous for evaluation purposes, as the correctness of model-generated answers can be straightforwardly determined by checking the final numerical value. This approach is in line with other benchmarks like GSM8k, SciBench, and MMMU. Besides calculations, our benchmark diversifies with other free-response types like Yes-or-No and fill-in-the-blank questions. These formats not only broaden the range of question types but also maintain ease of evaluation through exact matching. Given these characteristics, accuracy will be the primary metric for assessing performance on our benchmark.

In terms of data features, each problem was thoroughly reviewed by annotators to ensure it aligned with the intended high school and pre-college difficulty level. Moreover, every problem is accompanied by a clear explanation of the answer and is tagged with the main knowledge point from predefined knowledge sets. These annotations and categorizations have been verified by domain experts, ensuring that each problem accurately reflects the intended educational content and difficulty.

\section{Experimental Examination of SceMQA}
To determine whether our SceMQA dataset can serve as a benchmark for assessing current Multimodal Large Language Models (MLLMs), we test the latest MLLMs against it, reporting their performance in terms of answer accuracy. Additionally, we examine incorrect responses to identify the current MLLMs' limitations and demonstrate the value of our benchmark in exploring these limitations.

\subsection{Experimental Settings}
We choose InstructBLIP \cite{dai2023instructblip}, MiniGPT4 \cite{zhu2023minigpt} and LLaVa \cite{liu2023visual} as the open-source MLLM solvers for SceMQA. As for close-source models, we focus on three of the most representative MLLMs currently available: Google Bard, Gemini Pro and GPT4-V. Furthermore, we test GPT4-V and Gemini Pro under three distinct settings: zero-shot, few-shot, and text-only. In the zero-shot setting, the models are provided with the problem without any prior examples. The few-shot setting involves giving the models a small number of example problems and solutions to ``learn'' from, before attempting the new problems. We use hand-crafted text-only problems as examples since it is not flexible to insert multiple images in one API call. The text-only setting is a unique approach under zero-shot where only the textual content of the problem is provided to the model, without any accompanying images. All the prompts used in our experiments, along with detailed descriptions of each setting, are available for public view at \url{https://scemqa.github.io/}.

For the evaluation metric, we have chosen to use exact-match-based accuracy, which is consistent with several prior studies \cite{lu2023mathvista,yue2023mmmu} in this domain. This metric is particularly suitable for our benchmark as both the multiple-choice and free-response problems have definitive, singular correct answers. In the multiple-choice format, this involves selecting the correct option out of the presented choices. For the free-response format, it requires generating an accurate and precise answer, be it a numerical value, a yes/no response, or a specific term for fill-in-the-blank questions. Empirically we use rule-based answer exaction for multiple choice questions, and GPT4 as evaluators for free response questions.

\subsection{Accuracy for Solving  SceMQA}

The performance of examined MLLMs on SceMQA is presented in Table \ref{tab:combined_results}. Foremost, in all evaluated scenarios, the zero-shot GPT4-V consistently outperforms other models. Despite this, the challenge posed by the benchmark remains significant for even the most advanced MLLMs, including GPT4-V and Google Gemini. Notably, these models actually exhibit comparable performances in both SceMQA and a challenging college-level benchmark MMMU \cite{yue2023mmmu}. This parity shows the challenging nature of our benchmark and the necessity for further improving  MLLMs' reasoning capabilities.

It can be also observed that the performance of open-sourced models are significantly inferior to close-sourced ones. We have looked into the error cases and  found that the both instruction-following and reasoning abilities of open-sourced models are not very satisfactory, leaving a huge room for improvement.

\begin{figure*}
\centering 
\includegraphics[width=0.98\textwidth]{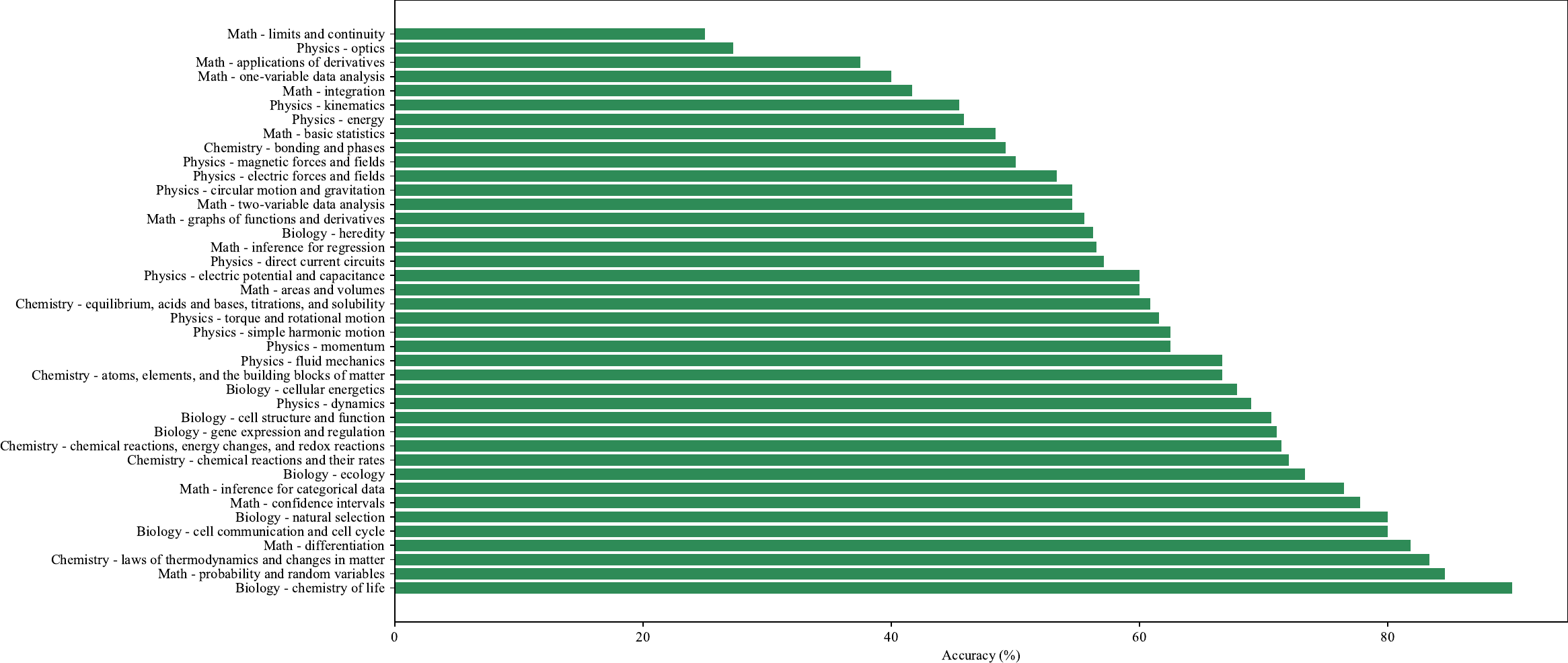} 
\caption{Accuracy distribution of GPT4-V on the knowledge points of SceMQA.} 
\label{fig:cate} 
\end{figure*}

\begin{figure*}
\centering 
\includegraphics[width=0.98\textwidth]{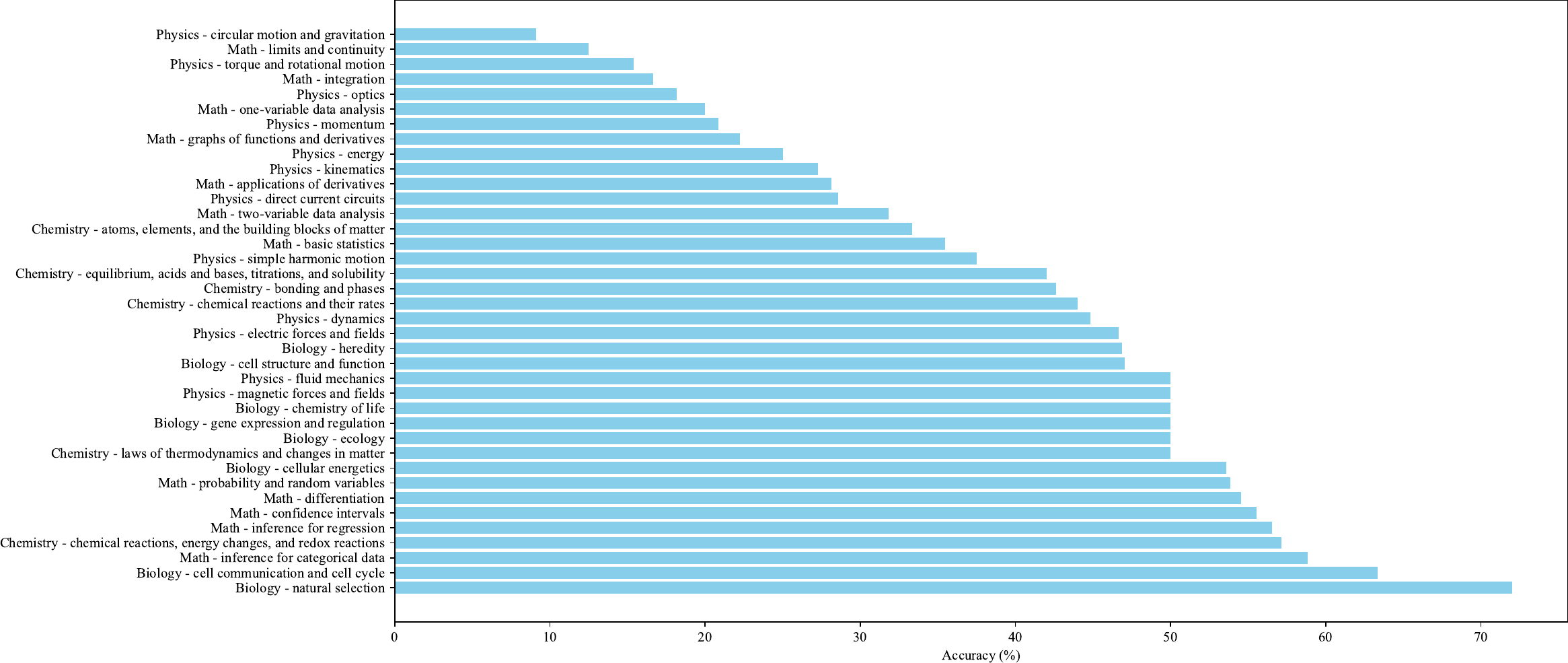} 
\caption{Accuracy distribution of Google Gemini on the knowledge points of  SceMQA.} 
\label{fig:cate_gemini} 
\end{figure*}

Additionally, in the few-shot setting, we noticed an intriguing trend: it underperforms the zero-shot setting. We hypothesize that the few-shot examples, while though providing guidance on scientific reasoning, do not enhance the models' ability to interpret scientific images. This could inadvertently lead the models to prioritize logical reasoning over critical image interpretation, leading to performance degradation. Also, we can see a significantly lower performance in the text-only setting. This result highlights the indispensability of visual information in solving the problems in our dataset.

Another notable finding is the variation in performance across different subjects. The models perform better in Chemistry and Biology compared to Math and Physics. We infer that this is because Math and Physics often require precise calculations for correct answers, while Chemistry and Biology tend to focus more on conceptual understanding. This pattern suggests that the integration of external computational tools, such as calculators or Python programs, might be beneficial in improving performance on our benchmark, particularly in subjects like Math and Physics.

\subsection{Accuracy across Knowledge Points}
In SceMQA, each problem is associated with  a specific knowledge point. The individual accuracy on those knowledge points can be found in Figure \ref{fig:cate} and \ref{fig:cate_gemini}. We can observe that the model generally performs better in chemistry and biology than in math and physics. Also, the worst-performed categories of knowledge points are generally related to image understanding (e.g., limits and continuity, optics) or calculation (e.g., one-variable data analysis, integration), which indicate the weaknesses of current MLLMs to some extent.

\begin{figure*}
\centering 
\includegraphics[width=1.01\textwidth]{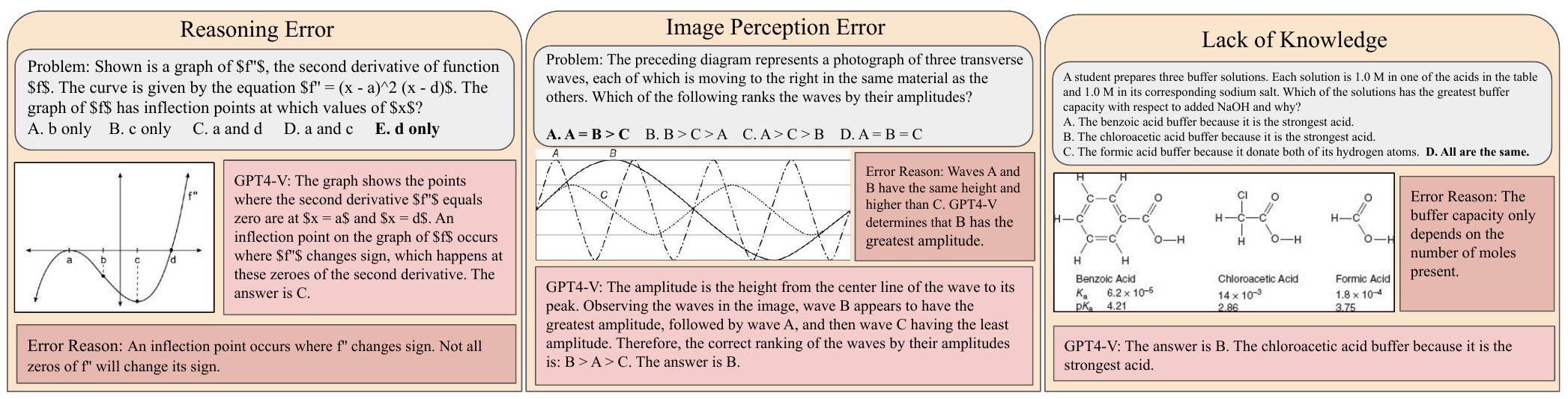} 
\caption{Example of errors made by GPT4-V on SceMQA.} 
\label{fig:error_example} 
\end{figure*}

\begin{figure}
\centering 
\includegraphics[width=0.38\textwidth]{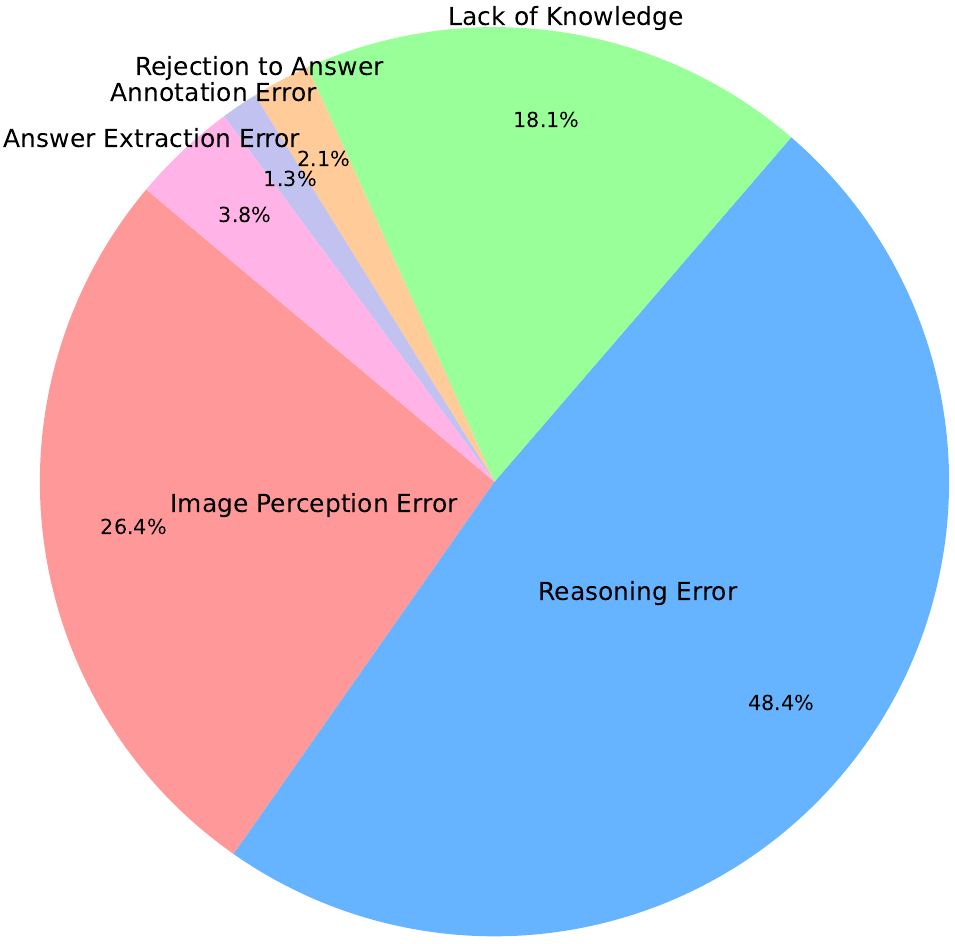} 
\caption{Distribution of GPT4-V's error types across 100 samples.} 
\label{fig:error} 
\end{figure}

\subsection{Error Analysis}
To delve deeper into the shortcomings of state-of-the-art MLLMs, we conducted a comprehensive error analysis. We randomly selected 150 instances of errors made by GPT4-V on the SceMQA dataset and enlisted two human experts for a detailed examination. These experts categorized each error into one of six categories: \emph{Image Perceptual Errors, Reasoning Errors, Lack of Knowledge, Rejection to Answer, Annotation Error}, and \emph{Answer Extraction Error}. The inter-rater reliability, assessed using the Kappa agreement score, was found to be greater than 0.5, indicating a moderate level of agreement between the annotators. We then averaged their annotations to determine the proportion of each error type, as depicted in Figure \ref{fig:error}. The top-3 error types are shown in Figure \ref{fig:error_example} and analyzed below:

\paragraph{Reasoning Error}
The most prevalent error type is categorized under \emph{Reasoning Error}. It occurs when the model correctly processes image-based information but fails to construct an accurate reasoning chain to arrive at the correct answer. Common mistakes include omitting necessary steps or making incorrect calculations. And we find these errors evenly spread in four subjects in SceMQA, underscoring the need for further development in the reasoning abilities of MLLMs. Drawing on insights from studies on LLMs, approaches such as prompting engineering \cite{wei2022chain} or supervised fine-tuning \cite{yu2023metamath,yue2023mammoth} might prove beneficial.

\paragraph{Image Perception Error}
This occurs when the model misinterprets visual information—such as incorrectly reading numbers or coordinates, or failing to differentiate between points in a geometric diagram. This type of error happens more often in the math subject because many math problems require precise diagram or table perception, which suggests that the image perception capabilities of current MLLMs require significant enhancement for precision and interpretation. Incorporation of external tools like OCR, as suggested in studies like \cite{liu2023hidden}, could potentially improve the model's understanding of visual content.

\paragraph{Lack of Knowledge}
This type of error arises when the model fails to correctly identify or apply relevant knowledge concepts, such as misusing formulas or misinterpreting theorems. These errors occur more in physics, chemistry and biology, which are indicative of gaps in the model's learned knowledge base, suggesting that enriching the training datasets of foundation models with diverse and domain-specific knowledge is essential to enhance their expertise in those domains.

\paragraph{Rejection to Answer and Annotation Error}
Interestingly, a smaller portion of errors were categorized as \emph{Rejection to Answer} and \emph{Annotation Error}. \emph{Rejection to Answer} occurs when the model refuses to provide an answer, possibly due to uncertainty or inability to comprehend the query. \emph{Annotation Error}, on the other hand, arises from inaccuracies or inconsistencies in the dataset's annotations, leading to confusion for the model. These categories highlight the importance of robust dataset design and also the need for models to handle ambiguous or complex instructions and questions effectively.

Through this detailed error analysis, we have identified specific patterns and weaknesses of MLLMs' performance on scientific problems. These findings provide valuable insights and directions for future research aimed at enhancing the capabilities of MLLMs. Addressing these identified issues could lead to significant improvements in the application of MLLMs in educational and research contexts, particularly in the domain of science.

\section{Conclusion}

In this paper, we introduced SceMQA, a novel multimodal question answering dataset tailored for the college entrance level, encompassing key scientific subjects: mathematics, physics, chemistry, and biology. A standout feature of SceMQA is its high annotation granularity, with over 90\% problems accompanied by detailed explanations and classified under specific knowledge points. Our experimental findings using state-of-the-art models like GPT4-V and Google Gemini highlight significant potential for further improvements in AI systems.

\bibliography{custom}

\appendix



\end{document}